\documentclass[journal]{IEEEtran}
\usepackage{framed,multirow}
\usepackage{amssymb}
\usepackage{latexsym}
\usepackage{amsmath,graphicx}
\ifCLASSINFOpdf

\else

\fi

\begin{document}

\title{SynthRAR: Ring Artifacts Reduction in CT with Unrolled Network and Synthetic Data Training}
\author{Hongxu Yang, Levente Lippenszky, Edina Timko, Gopal Avinash
\thanks{Hongxu Yang, Levente Lippenszky, Edina Timko, and Gopal Avinash are with Science\&Technology Organization, GE HealthCare (email: Hongxu.Yang@gehealthcare.com).}}

% make the title area
\maketitle

% As a general rule, do not put math, special symbols or citations
% in the abstract or keywords.

\begin{abstract}
Defective and inconsistent responses in CT detectors can cause ring and streak artifacts in the reconstructed images, making them unusable for clinical purposes. In recent years, several ring artifact reduction solutions have been proposed in the image domain or in the sinogram domain using supervised deep learning methods. However, these methods require dedicated datasets for training, leading to a high data collection cost. Furthermore, existing approaches focus exclusively on either image-space or sinogram-space correction, neglecting the intrinsic correlations from the forward operation of the CT geometry. Based on the theoretical analysis of non-ideal CT detector responses, the RAR problem is reformulated as an inverse problem by using an unrolled network, which considers non-ideal response together with linear forward-projection with CT geometry. Additionally, the intrinsic correlations of ring artifacts between the sinogram and image domains are leveraged through synthetic data derived from natural images, enabling the trained model to correct artifacts without requiring real-world clinical data. Extensive evaluations on diverse scanning geometries and anatomical regions demonstrate that the model trained on synthetic data consistently outperforms existing state-of-the-art methods.
\end{abstract}
% Note that keywords are not normally used for peerreview papers.

\IEEEpeerreviewmaketitle

\section{Introduction}
\begin{figure}[htbp]
\centering{\includegraphics[width=8cm]{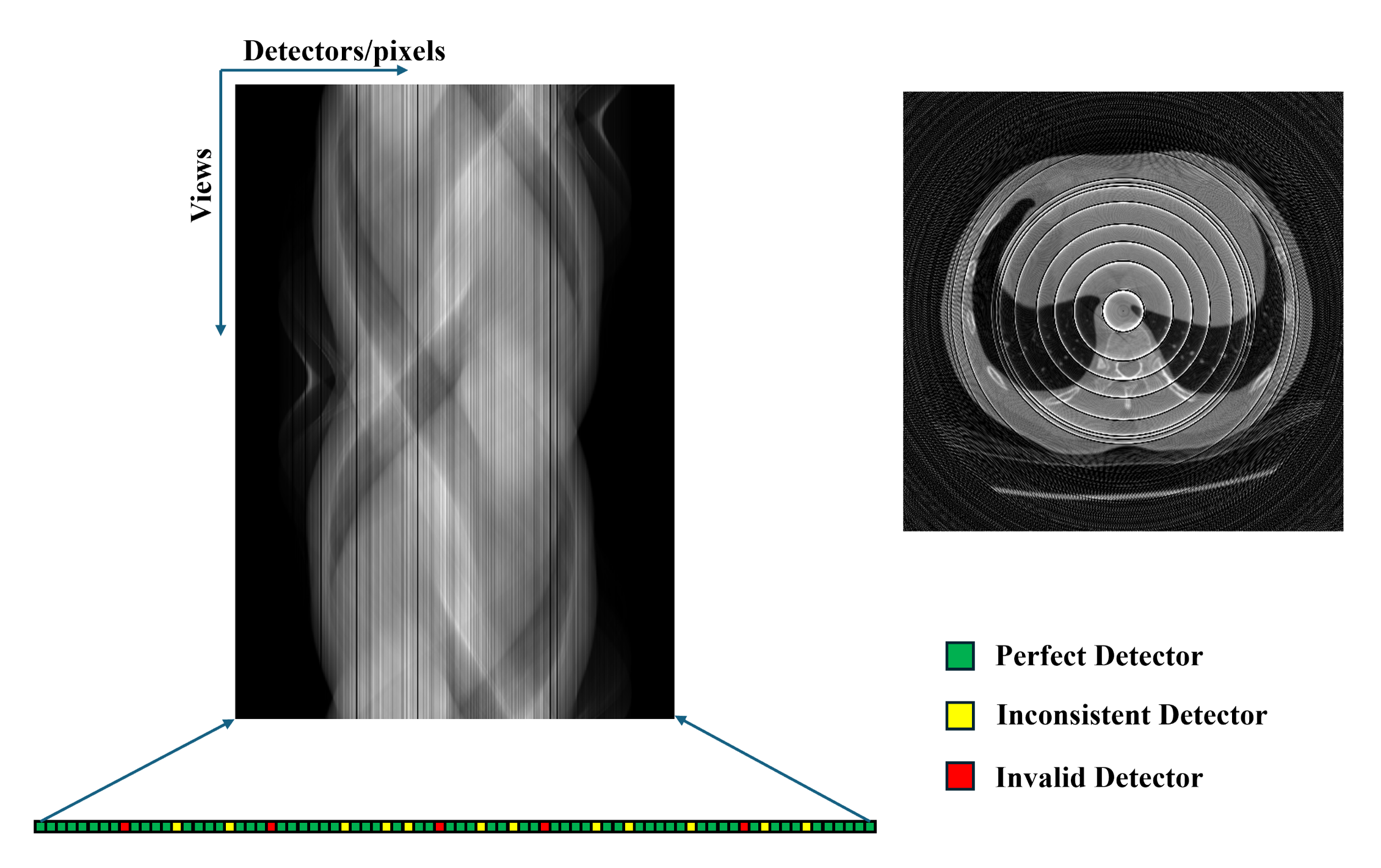}}
\caption{Illustration of non-ideal responses in sinogram data and corresponding reconstructed CT images, which has ring and streak artifacts due to inconsistent detector and invalid detector responses.}
\label{illustration}
\end{figure}

\IEEEPARstart{C}{omputed} Tomography (CT) is a key imaging modality in clinical workflows, offering rapid acquisitions and high-resolution visualization of patient anatomy, enabling disease diagnosis and intervention guidance. As medical imaging needs high-quality images with minimal artifacts, detector imperfections that lower image quality remain a major challenge that requires careful correction. Non-ideal detector pixels can result in missing or wrong signals in the sinogram and are commonly referred to as low performing pixels (LPP)~\cite{patil2022deep}. The most common artifacts caused by LPP are ring and streak artifacts. The severity of artifacts depends on the LPP position, i.e., greater artifacts occurring when LPP are near the iso-center, which may lead to an unusable image by traditional linear reconstruction algorithms, such as filtered back-projection (FBP)~\cite{patil2022deep}. Example images of sinogram data and FBP-reconstructed CT image are shown in Fig.~\ref{illustration}. The LPP pixels can be categorized into two categories: (1) those with completely missing responses, indicating fully defective detectors; (2) those with responses deviating from the ideal values, such as 10\% lower energy count than expected. In common practice, the former cases can be easily addressed by a calibration step. Solutions for fully defective detectors can be a complete detector module replacement, but it comes with high service costs and prolongs the waiting time for users. In contrast, the detectors exhibiting wrong responses are difficult to detect and require significant efforts in both software and hardware engineering, and potentially the replacement of the whole scanner. 

Recent studies have introduced various algorithm-based approaches to correct artifacts caused by LPP, such as image processing techniques or deep learning (DL) methods. Before the advancement of DL, interpolation-based techniques were commonly employed and demonstrated reasonable performance for fully invalid measurements (IM)~\cite{abdoli2011reduction,salehjahromi2019directional}. However, these approaches fail to address inconsistent response (IR) due to the difficulty in accurately quantifying detector deviations. More recently, several DL methods have been developed to perform RAR in the image or sinogram domain. DeepRAR~\cite{trapp2022deeprar} was proposed to correct the ring artifact in CT image using UNet~\cite{cciccek20163d}. Other image-based DL models were also considered for image restoration, such as CNN-based NAFNet~\cite{chen2022simple, wu2024unsupervised} and transformer-based AST~\cite{zhou2024adapt}. Alternatively, learning-based correction models for sinograms have also been proposed, such as Conjugate-CNN~\cite{patil2022deep} with supervised learning, SinoRAR~\cite{shi2025ring} and Riner~\cite{wu2024unsupervised} employing implicit neural representation method (INR). Most deep learning methods have shown promising performance, but still suffer from common limitations of supervised learning: high data collection costs and limited generalization. Although INR-based methods can alleviate the data requirements for model training, their high computational cost during test-time optimizations remains a key bottleneck for clinical applications. 

We propose SynthRAR, a novel supervised method for ring artifact reduction trained on synthetic data. Unlike conventional supervised models trained on real clinical data, synthetic data are utilized to incorporate the root causes of non-ideal responses across both detector and image domains. With training pairs generated for supervised learning, the ring artifacts are corrected through a dual-domain unrolled network, which iteratively corrects the artifacts in detection and image spaces. Based on the theoretical analysis of two types of low-performing pixels---invalid and inconsistent---we introduce two response estimation networks with an unrolled architecture. These networks are integrated into the linear forward projection of CT geometry, enabling an effective improvement of image quality. The proposed SynthRAR was evaluated on four datasets for 2D fan-beam CT on varying scanning geometries. The experimental results indicate that SynthRAR consistently outperforms existing SOTA solutions. More importantly, the proposed method demonstrates superior generalization across diverse anatomies and scanning configurations compared to traditional supervised learning approaches using real-world clinical data.

The main contributions of this paper are threefold. First, we propose a natural image-based synthetic data generation strategy for RAR, which is designed to generate training pairs. Second, we perform RAR by a dual-domain unrolled model, reformulating the LPP problem within the linear forward projection operation. Lastly, we experimented on a public CT dataset to demonstrate that the proposed method outperforms existing approaches trained on real CT data by a large margin. We have extended our preliminary conference study~\cite{yang2026low} by estimating both invalid and inconsistent detectors. Moreover, the this paper includes more detailed formulation of the considered challenges with newly designed modules, which is able to estimate the detector responses rather than assuming the knowledge of invalid detectors. Moreover, the proposed method is extensively compared to several different datasets. Therefore, this paper has more than 70\% content than our conference paper. The remainder of this paper describes the details of our proposed method in Section \ref{methods}. The experiments, results and discussions are discussed in Section \ref{experiments} and \ref{results}, respectively. Section \ref{conclusions} provides conclusions for the paper.

\section{Related Work}\label{literature}
\subsection{State-of-the-art RAR}
Ring artifact reduction is one of the most challenging and expensive problems in CT imaging~\cite{chen2024research}. Conventional ring artifact solution is to replace all possible detectors from the scanner, which comes with high service costs and extends user waiting times. To mitigate the service cost, researchers and engineers have explored software-based solutions. Interpolation and handcrafted filtering techniques were commonly studied before the era of deep learning and achieved reasonable performance~\cite{abdoli2011reduction,salehjahromi2019directional}. But these methods are constrained by insufficient priors, as manual design often relies on high-level abstract assumptions. In recent years, data-driven deep learning solutions have achieved promising results for RAR challenges using large-scale paired clinical datasets~\cite{trapp2022deeprar, patil2022deep}. Nevertheless, these end-to-end models are data-dependent and exhibit poor generalization when applied to out-of-domain cases. For example, a model trained on chest CTs will perform poorly when applied to abdominal scans. More recently, the popularity of implicit  neural representation (INR) increased and several unsupervised methods were proposed to perform RAR during testing time~\cite{shi2025ring,wu2024unsupervised}. Nevertheless, due to limited physical prior information in the observed data and the high computational cost of post-scanning optimization, these solutions remain impractical for clinical practice. 

\subsection{Inverse Problem by Deep Learning}
Inverse problems are widely studied in the medical imaging domain, based on the fundamental characteristics  of data acquisition by the scanner~\cite{aggarwal2018modl}. One of the most studied topics are related to accelerating image acquisition leveraging compressed-sensing concept to capture sparse data through a linear transformation, such as encoding matrix in k-space for MRI~\cite{hu2021self}, sparse-view scanning for CT~\cite{sun2025efficient}, and kernel synthesis in image domain~\cite{aggarwal2025display}. These methods are typically implemented by deep-learning approaches combined with prior-knowledge operations, such as k-space sampling patterns, scanning views, or signal-of-interest frequency selection. More recently, INR-based solutions have been proposed to correct LPP artifacts through multi-parameter estimation, based on the physical principles behind CT artifact formation~\cite{wu2024unsupervised}. Inspired by the multi-parameter inverse problem~\cite{wu2024unsupervised}, we extend the standard unrolled model by incorporating detailed physics-based mechanisms of CT artifact formation to enhance ring artifact reduction.

\subsection{Synthetic Data for Deep Learning}
Synthetic data-based deep learning models have gained popularity in recent years, as a mitigation for the lack of training image and ground truth pairs across various applications. An example of this approach is proposed for synthetic image-based tissue segmentation utilizing an analytical approach~\cite{billot2023synthseg}. Similarly, image reconstruction and acceleration methods~\cite{hu2021self,sun2025efficient} are frequently used to generate synthetic training data by embedding physical priors and introducing specifically designed artifact and pattern, therefore reducing the cost of collecting real-world data. Synthetic noise is another widely used strategy in denoising tasks, which has achieved promising and mature results in practical application~\cite{kang2017deep}. Despite their advantages, these methods synthesize task-specific artifacts on real clinical images, inevitably introducing bias while requiring substantial efforts of data collection. To address these challenges, we consider the simulation of ring artifacts under the guidance of physical principles on random natural images following a standard CT scanning protocol. This approach eliminates training data collection efforts and mitigates out-of-distribution challenges.

\section{Methods}\label{methods}
\subsection{Theoretical Analysis of CT Ring Artifacts}
A standard CT detector operates as an energy-integrating detector. In a CT scanner, the source-detector pair rotates around the gantry by $360^\circ$, and the information received by the detector channels corresponds to individual scanning views~\cite{patil2022deep}. For a given view, the readout from each detector channel is obtained by the line integral of the X-ray attenuation through the materials along the path of an X-ray beam with energy $E$. Based on Lambert-Beer's law, the readout data is expressed as 
\begin{equation}
I(E)=N\eta\exp(-\int_{0}^{l_1}\mu(E,\vec{l})dl)
\end{equation}
where $N$ is the total number of incidental photons, $\eta$ is the detector response function w.r.t. photon energy, $\mu(E,\vec{l})$ is the energy attenuation coefficient at energy level $E$, vector $\vec{l}$ is location and ${l_1}$ is the propagation path length. 

As for LPP, the received signal can vary due to differences in the response functions. When a detector pixel fails or reaches a near-dead state, the response function $\eta$ is approaching zero. Unlike completely dead pixels, the open circuit may produce unstable values due to diode variations, potentially affecting neighborhood detectors, which will generate inconsistent signal response. 

Assuming the ideal detector, where $\eta=1$, the actual CT measurement sinogram data can be expressed as
\begin{equation}
S=-\ln{\frac{I(E)}{{I_0}}}=-\int_{0}^{l_1}\mu(E,\vec{l})dl
\end{equation}
With the above functions, the CT image can be reconstructed from ideal measurement data using the standard filtered back-projection (FBP) algorithm~\cite{wu2024unsupervised} for number of photons by X-ray source $I_0$. 

When $\eta$ is non-ideal and inconsistent across the detector due to hardware failure, the actual measurement is defined below when $\eta>0$ 
\begin{equation}
S=-\ln{\eta} + S
\end{equation}
From the definitions above, invalid measurements (IM) occur when $\eta=0$, whereas inconsistent responses (IR) arise when $\eta>0$ and different detectors exhibit various values of $\eta$~\cite{wu2024unsupervised}. With these two cases considered in the observed data, the corresponding image reconstructed by FBP includes ring and streak artifacts if no correction is applied. 

\subsection{Problem Formulation}
Existing methods either correct artifacts in the image domain as standard restoration tasks~\cite{trapp2022deeprar}, or address missing and inconsistent data issues in the sinogram domain as an interpolation and filtering task~\cite{patil2022deep,shi2025ring}. Although Riner~\cite{wu2024unsupervised} proposed a joint dual-domain method to correct the artifacts, its test time optimization is time-consuming and requires intensive calculations for each grid point. In the proposed method, the RAR task is formulated as an optimization problem expressed as 
\begin{equation}
\arg \min_{x} \frac{1}{2}||Ax - y||_2^2 + \lambda R(x)
\end{equation}
where $x$ is the desired image, $y$ is the observed sinogram data with artifacts, $A$ is a linear forward projection, such as fan-beam, parallel-beam or cone-beam geometry. The function $R(x)$ denotes the regularization with the weight parameter $\lambda$. The objective of CT artifact correction is to recover the desired image $x$ from its measurement $y$. 

Specifically, linear forward projection $A$ is a simple projection matrix for an ideal CT scanner, but can be generalized to non-ideal by considering the above IM and IR cases, which is formulated as
\begin{equation}
\begin{split}
\tilde{A}x &= {H}_{IM}(-\ln{{H}_{IR}} + Ax)\\
&=-{H}_{IM}\ln{{H}_{IR}}+{H}_{IM}Ax
\end{split}
\end{equation}
where ${{H}_{IR}}$ is the detector response when detectors have inconsistent response (IR), while ${H}_{IM}$ is a binary mask for invalid measurement (IM) when detectors have no response. IR and IM operations are point-wise processes on each data point in the sinogram $Ax$. It can be observed, that the non-ideal detectors are essentially impacted by a sampling mask ${H}_{IM}$ and background additive noise ${H}_{IM}\ln{{H}_{IR}}$.

To solve the above optimization problem, the popular first-order proximal iterative shrinkage thresholding algorithm (ISTA)~\cite{beck2009fast} is adopted for the linear inverse problem. To incorporate data-driven priors, ISTA-Net~\cite{zhang2018ista} is considered in our method, for its simplicity and effectiveness. Specifically, ISTA-Net assumes $R(x)$ transforms coefficients of $x$ with respect to some transform $R(\cdot)$, and the sparsity of the vector $R(\cdot)$ is encouraged by the $L_1$ norm with weight being the (generally pre-defined) regularization parameter~\cite{zhang2018ista}. The ISTA-Net iteratively applies two steps to optimize the problem~\cite{pezzotti2020adaptive}:
\begin{equation}
r^{(k)}=x^{(k-1)} - \rho {A}^{-1}({H}_{IM}Ax^{(k-1)}-{H}_{IM}\ln{{H}_{IR}}- y)
\end{equation}
\begin{equation}
x^{(k)}=\hat{G}(\operatorname{soft}(G(r^{(k)}), \theta))
\end{equation}
where $k$ denotes the iteration index, $\rho$ is the step size, $G(\cdot)$ and $\hat{G}(\cdot)$ are CNN-based image trainable operations, with $\hat{\cdot}$ denoting the symmetrical operation of the structure. The parameter $\theta$ represents the shrinkage threshold for the soft thresholding operation $\operatorname{soft}$. Operation ${A}$ is the forward projection of the scanning protocol with integration of non-ideal detector responses, while ${A}^{-1}$ is the reverse of forward projection. All parameters in the ISTA-Net can be optimized by end-to-end training~\cite{zhang2018ista}. Specifically, to measure $\eta$ and $\eta_0$, we follow the suggestion of Riner~\cite{wu2024unsupervised} where two sinogram-space convolutional neural networks (CNNs) are jointly trained with ISTA-Net to estimate IM and IR separately. 

\subsection{Model Design}
\begin{figure}[htbp]
\centering{\includegraphics[width=8cm]{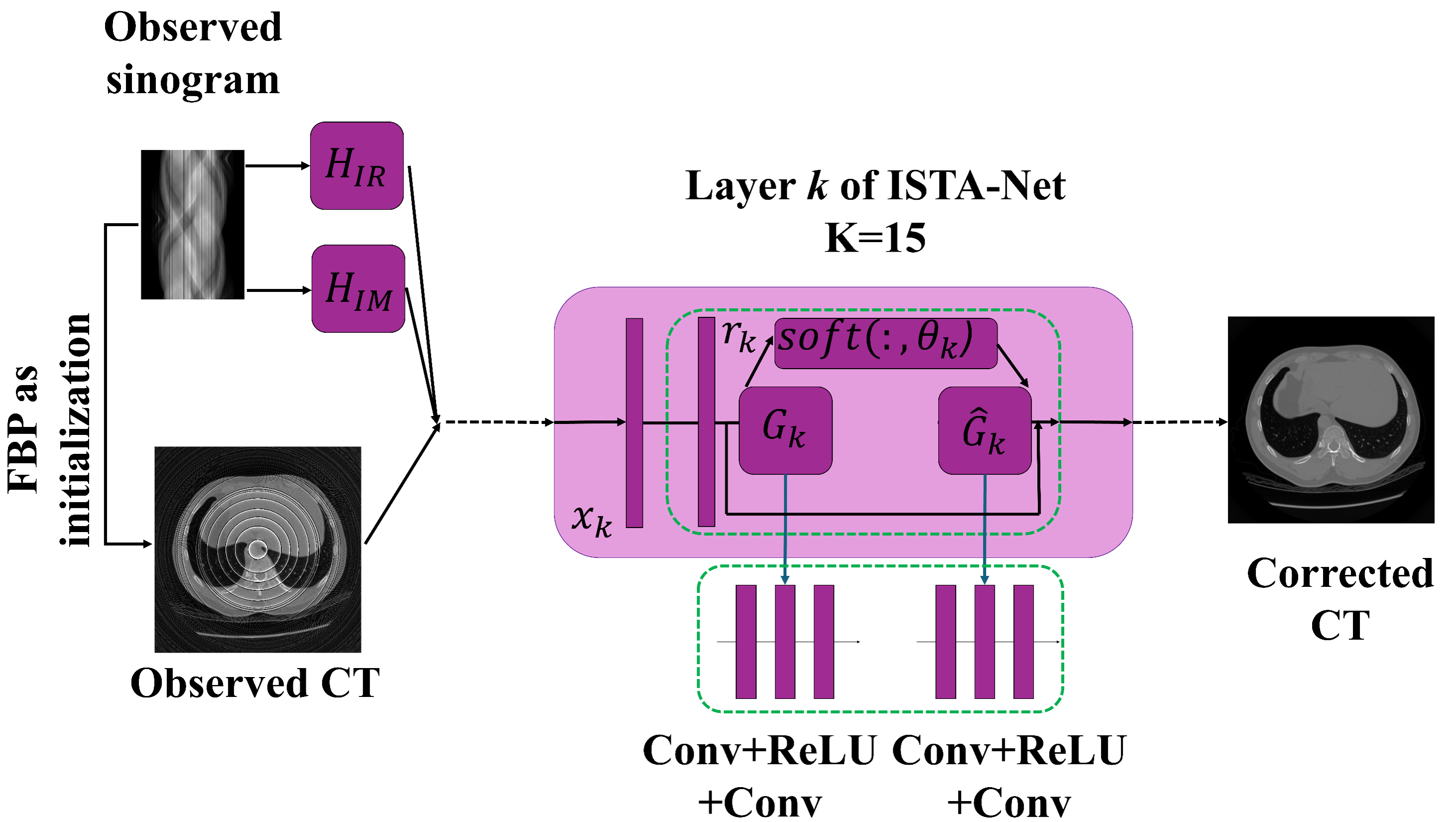}}
\caption{Illustration of the proposed SynthRAR model, which utilizes the basic ISTA-Net with modified components for IR and IM estimations. The model performs ring artifact correction through K iterative updates, progressively refining the reconstruction at each iteration.}
\label{overall_model}
\end{figure}

Building on the preceding formulation, we propose SynthRAR, a physics-informed ring artifact reduction model to mitigate artifacts. The method leverages three complementary strategies to accomplish the RAR task: (1) integrating LPP estimation networks into the ISTA-Net backbone architecture, (2) applying a refined forward projection formula $\tilde{A}x = (-\ln{{H}_{IR}} + Ax)\times{H}_{IM}$, and (3) generating synthetic training data with a designed root cause of LPP. The overall model architecture is shown in Fig.~\ref{overall_model}.

\subsubsection{ISTA-Net}
ISTA-Net is an unrolled design of ISTA algorithm to iteratively estimate dynamic gradient descent module and dynamic proximal mapping module for K iterations, corresponding to Eqn. (6) and (7), respectively. In our implementation we follow the original  ISTA-Net design for these modules with K=15, as depicted in Fig.~\ref{overall_model}. More detailed implementation can be find in the original paper~\cite{zhang2018ista,hu2021self}. 
\subsubsection{Modified Forward Projection}
\begin{figure}[htbp]
\centering{\includegraphics[width=8cm]{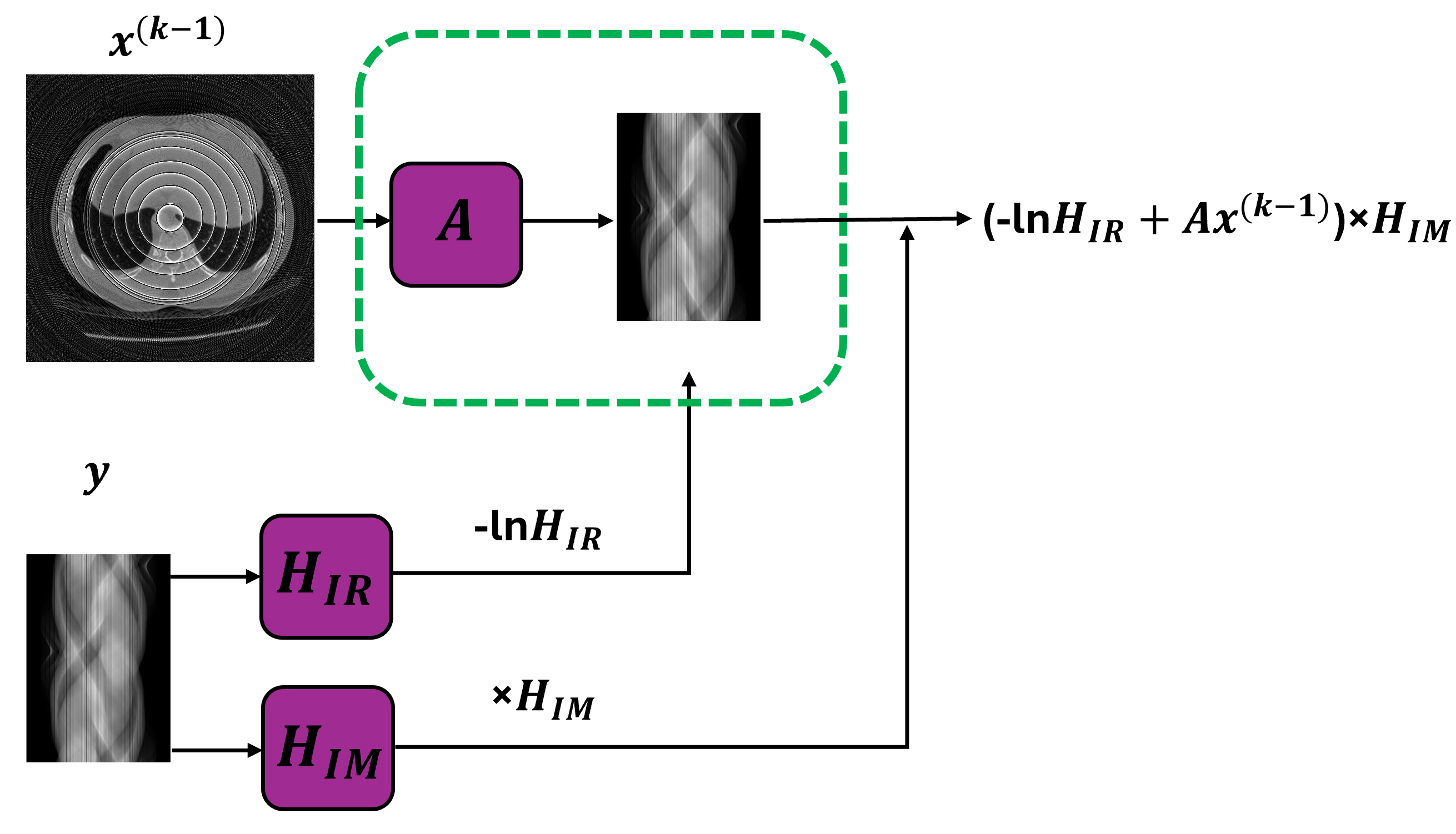}}
\caption{Illustration of modified forward projection step. The input CT at iteration k is transformed to sinogram with known CT geometry. In addition, the original sinogram is processed by two individual networks to estimate IR and IM measurements for function $\tilde{A}x = (-\ln{{H}_{IR}} + Ax)\times{H}_{IM}$.}
\label{forward_projection}
\end{figure}
As shown in Eqn.(5), the standard forward projection is modified to approximate the LPP cases by incorporating inconsistent detector responses (IR) and fully invalid measurement (IM). From Eqn.(3) and (5), the IR component can be interpreted as introducing minor scaling factors into the observed sinogram data  as a consequence from its normalization to original energy counts as dictated by Lambert-Beer's law. In contrast, IM detector responses are easier to model as zero-valued signals. As shown in Eqn.(5), these operations preserve the linearity assumptions of ISTA solver, allowing the original forward operator $A$ to be replaced by $\tilde{A}$ in the ISTA-Net implementation. The designed module is shown in Fig.~\ref{forward_projection}. To estimate $H_{IR}$ and $H_{IM}$, two lightweight CNNs are designed to predict detector-level response vectors across all scanning views, leveraging the fixed defective properties of detector positions and their correlations in the observed sinogram. To satisfy the definition of IR and IM, the network outputs are projected by sigmoid operation with scaling factors, ranging from [0.75, 1.25] for IR and [0,1] for IM, as suggested by Riner~\cite{shi2025ring}.

\begin{figure}[htbp]
\centering{\includegraphics[width=8cm]{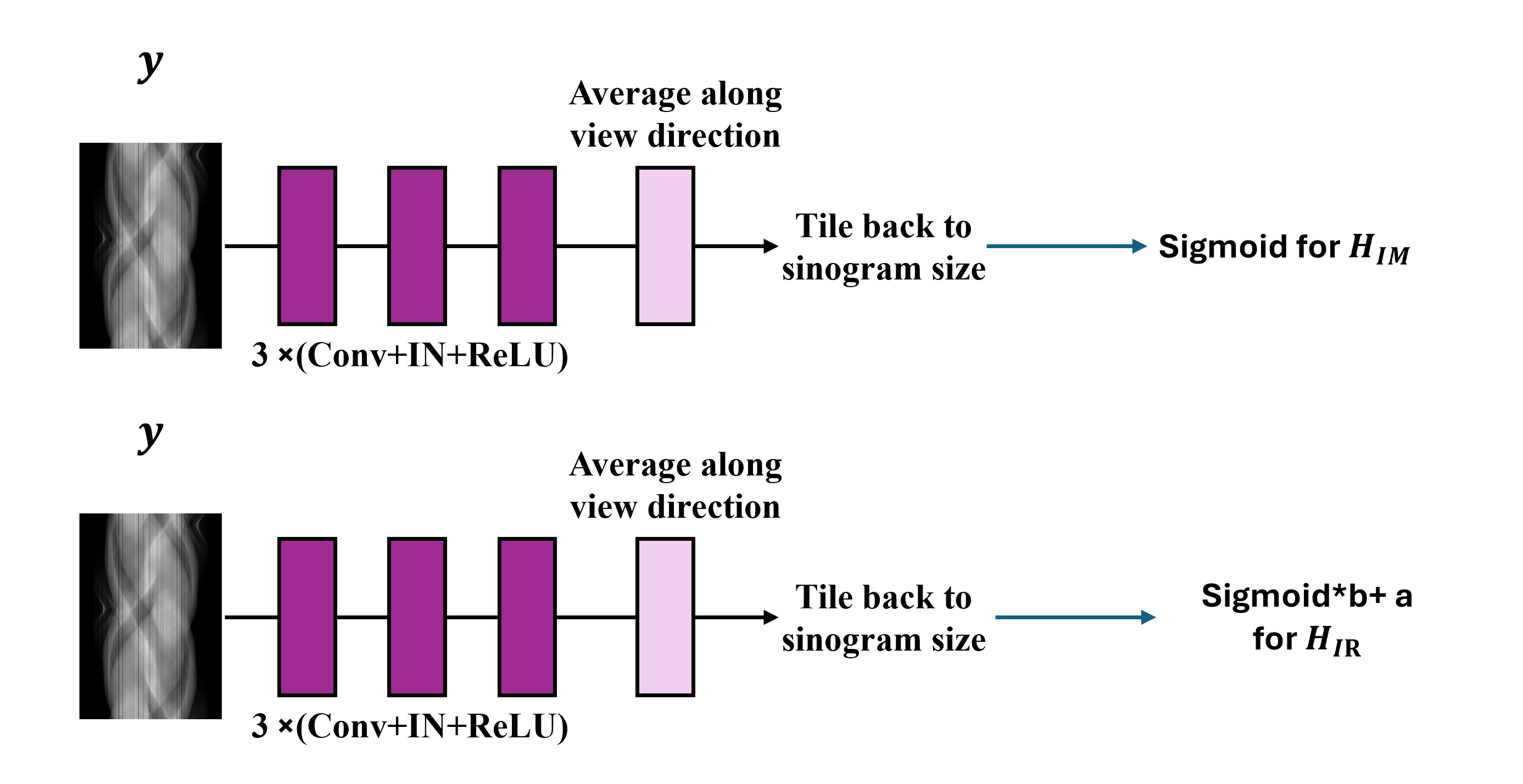}}
\caption{Illustration of networks to estimate IR and IM from sinogram. The IR output is processed by Sigmoid and rescaled to the common definition range of inconsistent responses (a=0.75 and b=0.5 in our implementation). The IM output is processed by Sigmoid to predict binary mask of invalid detectors.}
\label{estimation_net}
\end{figure}

\subsubsection{Synthetic Data Design}
\begin{figure*}[htbp]
\centering{\includegraphics[width=12cm]{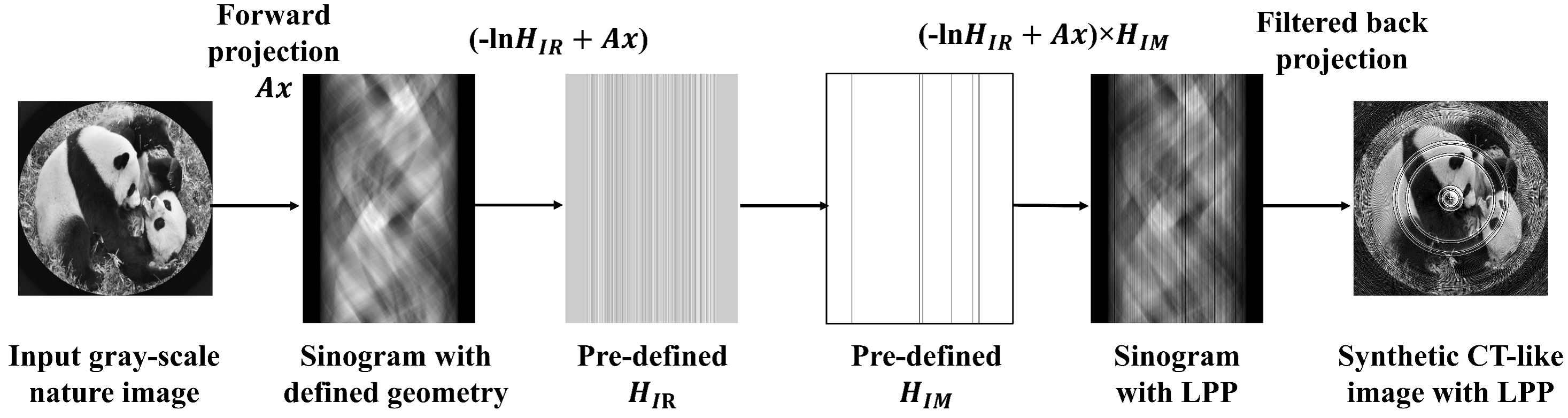}}
\caption{Illustration of synthetic CT-like training data generation based on the definitions of forward projection, IR and IM. The natural gray-scale image is masked by a circle to simulate the CT appearance with re-scaled intensity value to random range 0.5-0.7 for linear attenuation coefficient $\mu$.}
\label{synthetic_ct}
\end{figure*}
Commonly used image restoration approaches, such as CNN- or Transformer-based solutions, require image and ground truth (GT) pairs for training, which introduce data collection efforts and suffer from limited generalization. In addition, medical data collected under specific region-of-interest and scanner settings cannot be transferred to a different scanner protocol, creating additional requirements for data collection across different body parts, vendors and scanners. With the above formulation of IR and IM, the model will focus only on data correction with the guidance of transformation $(-\ln{{H}_{IR}} + Ax)\times{H}_{IM}$ in CT. Therefore, the model is expected to correct the artifacts in the dual-domain iteratively without considering the actual content of the image from any resources from the synthetic data.

In our settings, natural images were obtained from generic datasets and converted to grayscale. For a grayscale image, the maximum intensity is randomly scaled to a linear attenuation coefficient $\mu$ within the empirical range $[0.5, 0.7]$~\cite{zhang2018convolutional}. Then, with known predefined CT geometry, such as a fan-beam CT in this paper, the forward operation is applied to the converted grayscale image to obtain the sinogram data. To simulate non-ideal points on the detector, a random number of detector signals are zeroed out, while 75\% responses are randomly rescaled in range [75\%, 125\%], as suggested in~\cite{shi2025ring}. Finally, the corrupted sinogram is reconstructed as synthetic CT using a standard filtered back-projection algorithm with ramp filter. The corresponding IR and IM values are used to serve as the ground truth for model training. 

\subsubsection{Model Optimization}
With the designed model and training dataset, the proposed SynthRAR can be optimized by the following loss function 

\begin{equation}
\begin{split}
L(\theta_1,\theta_2,\theta_3) 
&= \frac{1}{N} \sum_{i=1}^N \Big( 
    MSE(x_i, f(\theta_1; y_i)) \\
&\quad + MSE(IR_i, g(\theta_2; y_i)) \\
&\quad + MSE(IM_i, h(\theta_3; y_i)) 
\Big) + \lambda L_{\text{cons}}
\end{split}
\end{equation}

where $N$ is the number of training cases, $i$ denotes the $i^{th}$ (image,ground truth) pair for $(y_i, x_i)$, parameter $IR_i$ and $IM_i$ are known ground truth for inconsistent response and invalid measurement on the detector given the CT scanning geometry. Parameter $\theta_1,\theta_2,\theta_3$ represent the CNN learnable parameters of backbone ISTA-Net $f(\theta_1; \cdot)$, IR estimation net $g(\theta_2; \cdot)$, and IM estimation net $h(\theta_3; \cdot)$. $MSE$ is mean squared error loss function. $\lambda L_{cons}$ is constraint loss of ISTA-Net~\cite{zhang2018ista,hu2021self}, which ensure the learned transform $\hat{G}$ and $G$ to satisfy the symmetry constraint~\cite{zhang2018ista,hu2021self}. In the end, parameter $\lambda$ is a weight factor for the constraint loss.
\section{Experiments}\label{experiments}
\subsection{Dataset}
Two types of datasets were considered in the experiments. The synthetic dataset was used to train SynthRAR. Additionally, four public CT datasets were employed to train and evaluate performance against other SOTA methods for comparison.
\subsubsection{Synthetic Dataset}
SynthRAR was trained on synthetic images designed to mimic CT characteristics. This dataset was generated using approximately 5000 color images from the public ILSVRC2017 dataset~\cite{ILSVRC15}. The images were converted to grayscale and resized to $512\times512$ pixels and then transformed into the sinogram space with a predefined geometry based on TorchRadon~\cite{ronchetti2020torchradon}. Specifically, we defined a fan-beam CT scanner with six different settings, which were randomly assigned during synthetic data generation. The detailed fan-beam CT geometries are shown in Table~\ref{geometry}. To simulate IR detectors, 75\% points of the total detector responses were randomly selected as inconsistent with response factors $\eta$ uniformly sampled within the range of [75\%, 125\%]. To simulate IM, 2\% of the total number of detectors were randomly selected and zeroed-out to represent detector failures. These parameters were selected based on the literature to show the capability of ring artifact correction~\cite{patil2022deep,wu2024unsupervised}. For each image in the ILSVRC2017 dataset, 10 random augmentation were generated, resulting in approximately 50k training samples for SynthRAR.

\begin{table*}[htbp]
\centering
\caption{Detailed 2D Fan Bean CT scanning geometry of synthetic CT based on ILSVRC2017 dataset. The same settings are also used for real CT data for validation.}
\label{geometry}
\begin{tabular}{lccccccc}
\hline
Geometry No.           & 1& 2& 3 & 4& 5& 6 & LDCT\\ \hline
Image Size           & $512\times512$& $512\times512$& $512\times512$ & $512\times512$& $512\times512$& $512\times512$& $512\times512$\\ 
Pixel Size (mm$^2$)          & $1\times1$& $1\times1$& $1\times1$ & $1\times1$& $1\times1$& $1\times1$&$0.6641\times0.6641$\\
View Range ($^{\circ}$)           & [0, 2$\pi$) & [0, 2$\pi$) & [0, 2$\pi$)& [0, 2$\pi$)& [0, 2$\pi$) & [0, 2$\pi$)& [0, 2$\pi$)\\
Number of Detectors           & 681& 758& 641 & 880& 801& 900 & 736\\ 
Detector Spacing (mm)           & 2& 2& 2.133 & 2 & 2 & 1.727 & 1.2858\\ 
Number of Views           & 984& 1024& 720 & 840 &  900 & 1000 & 984\\ 
Distance from Source to Center (mm)           & 722& 1075& 750 & 1000& 950& 550& 595\\ 
Distance from Center to Detector (mm)           & 722& 1075& 850 & 1000& 950& 400 &490.6\\ 
\hline
\end{tabular}
\end{table*}

\subsubsection{In-distribution Real CT Dataset}
To evaluate SynthRAR on actual CT images and perform comparison with other SOTA methods, we considered two different public datasets: MMWHS CT dataset~\cite{zhuang2018multivariate}, and RibFrac dataset~\cite{yang2025deep}. As for MMWHS CT dataset~\cite{zhuang2018multivariate}, 20 volumes of training images were used to generate CT images with RAR artifacts for comparison with other supervised learning methods, while 40 volumes were held out for testing and validation. For each training volume, axial slices containing semantic masks from MMWHS~\cite{zhuang2018multivariate} were extracted for training without restriction, while the central axial slice for each testing volume was used as the evaluation image. This process resulted in around 50k CT slices for training and 40 CT slices for testing. The HU values are transformed to $\mu$ unit by using $\mu_{water}=0.268$ and $\mu_{air}=0$. The RAR simulation and the fan-beam geometries followed the synthetic data generation strategy described above. Regarding the RibFrac dataset~\cite{yang2025deep}, 40 volumes were used to generate the training dataset under the same strategy of MMWHS dataset, which generated around 130k training cases. In addition, 40 volumes were used to generate 40 testing images.
\subsubsection{Out-of-distribution Real CT Dataset}
To evaluate the generalization capability of the model on the out-of-distribution CT images, i.e., different distribution of scanning geometry and image-level HU values, two real-CT datasets were considered. First, seven volumes of size $512\times512$ pixels along the axial direction from the RIRE dataset~\cite{west1996comparison} were processed to generate $5\times6$ different test slices for generalization validation purposes (7 slices and 6 settings). This dataset was used to evaluate the distribution mismatch in HU values, as the brain CT would generate lower HU differences in soft tissues. Second, the LDCT dataset~\cite{xia2023physics} from a different scanning setup was considered to evaluate the SynthRAR performance on different scanner geometry. In our experiment, 40 images from the shared validation dataset were used for evaluation, with IR and IM detectors simulated as described above.

\subsection{Baselines and Metrics}
We compared SynthRAR to several SOTA image restoration solutions, including (1) the baseline filtered-back projection (FBP), (2) model-based algorithms Norm~\cite{rivers1998tutorial}, WaveFFT~\cite{munch2009stripe} and Super~\cite{vo2018superior}, (3) supervised restoration DL models, such as AST~\cite{zhou2024adapt}, DeepRAR~\cite{trapp2022deeprar}, NAFNet~\cite{chen2022simple}, and (4) most recent unsupervised Riner model with Implicit Neural Representation (INR) solution~\cite{wu2024unsupervised}. The compared DL models were trained on MMWHS and RibFrac images separately, while SynthRAR was trained on synthetic CT-like data. The evaluations were performed on MMWHS, RibFrac, RIRE and LDCT test images. The results are measured by Peak Signal-to-Noise Ratio (PSNR), Mean Absolute Error (MAE) and Structural Similarity Index Measure (SSIM) in HU values. 

\subsection{Implementation Details}
The ISTA-Net follows the original implementation, with $K = 15$ layers and 64 convolutional kernels. Similarly, the CNNs used for estimating IR and IM consists of three sequential blocks of Conv+IN+ReLU with 64 filter kernels. The parameter $\lambda$ in the loss function is selected to be 0.01 based on the original work. SynthRAR was trained using the Adam optimizer with mini-batch size of 1 for 500k steps on a single NVIDIA A100 GPU. The SOTA models were directly obtained from their official releases and integrated into the same training pipeline as SynthRAR to ensure a fair comparison. As for Riner~\cite{shi2025ring}, the scanning settings were modified according to our proposal to accommodate different experimental geometries.

\section{Results}\label{results}
\subsection{Performance comparison to SOTA methods on in-distribution images}
The ring artifacts reduction performance metrics are summarized in Table~\ref{supervised_performance}. The proposed SynthRAR achieved the best quantitative results across both datasets, outperforming SOTA supervised learning solutions, which were trained and tested in the same data domain (i.e., MMWHS–MMWHS and RibFrac–RibFrac pairs). The state-of-the-art Riner employing INR yielded the worst results and was also the most time-consuming, as it requires online optimization for each observed image. Compared to its original assumption of only two invalid measurement points, we considered 2\% of all detectors, corresponding to approximately 10–20 points. 

\begin{figure*}[htbp]
\centering{\includegraphics[width=12cm]{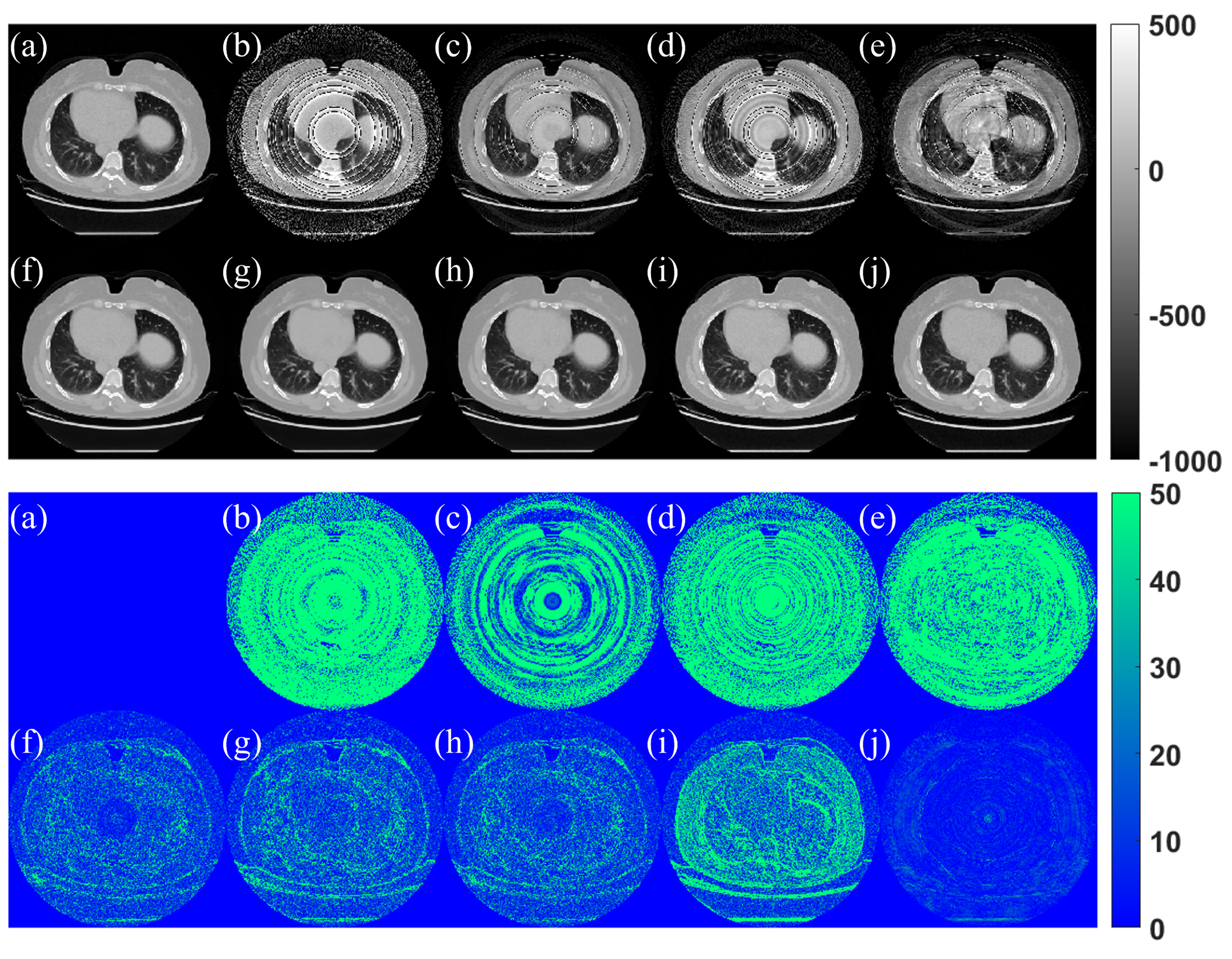}}
\caption{Qualitative comparison for RAR methods: (a) Ground truth, (b) FBP, (c) Norm, (d) WaveFFT, (e) Super, (f) AST, (g) DeepRAR (h) NAFNet, (i) Riner, and (j) SynthRAR. Top: images in HU value. Bottom: absolute HU value differences to ground truth.}
\label{ribfrac_compare_hu_self}
\end{figure*}

\begin{figure}[htbp]
\centering{\includegraphics[width=6cm]{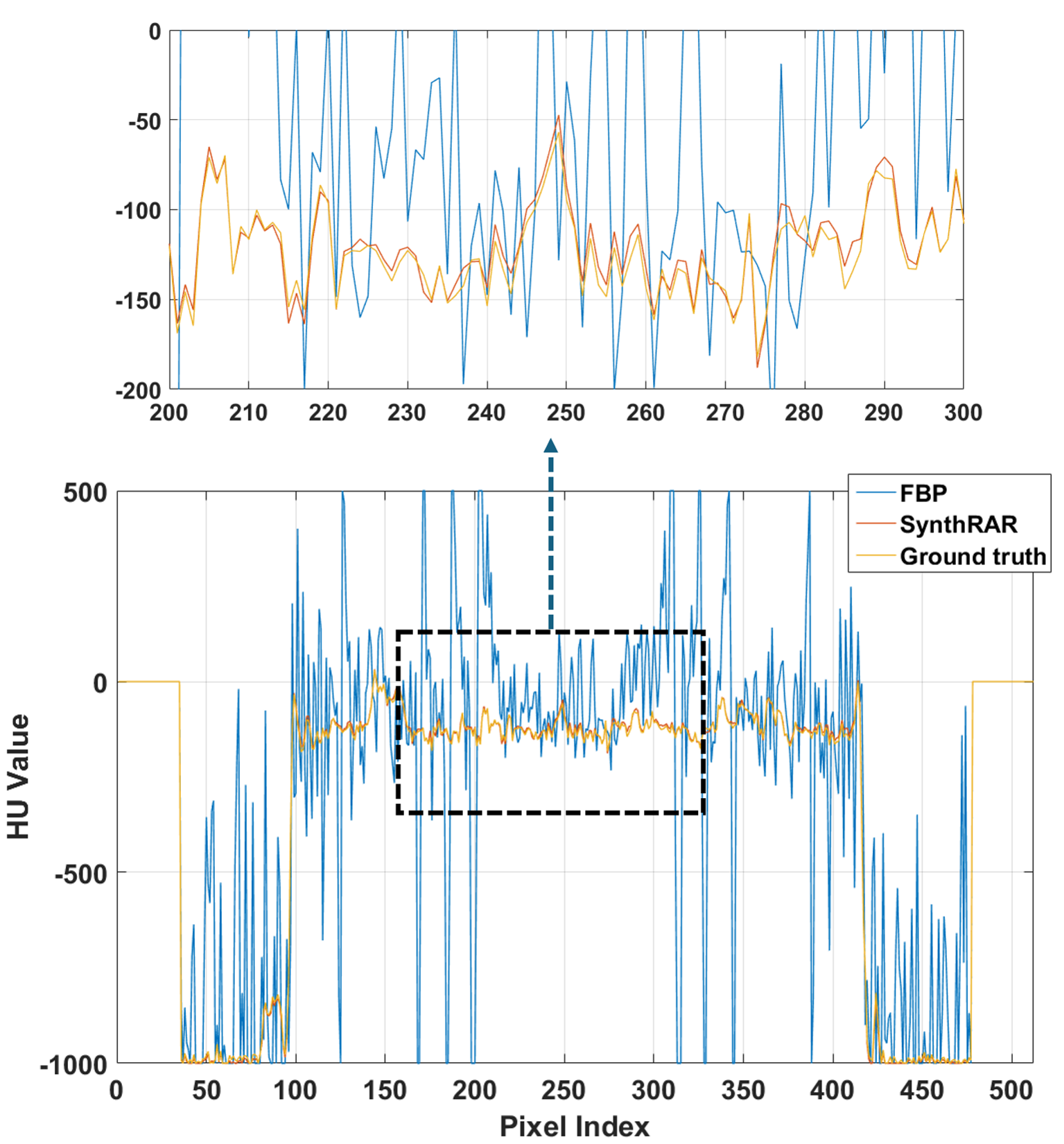}}
\caption{Profile analysis in image domain (row 128), corresponding to images from Fig.~\ref{ribfrac_compare_hu_self}. The enlarged region spans pixel indices between 200 and 300.}
\label{hu_rois}
\end{figure}

A key observation is that Riner introduced significant HU value shifts while preserving CT contextual information, which resulted in poor quantitative performance for CT quantification tasks. From qualitative results in Fig.~\ref{ribfrac_compare_hu_self}, all supervised methods trained on CT images, such as AST, DeepRAR and NAFNet can correct the artifacts but did not outperform SynthRAR. The detailed HU value difference profiling among FBP, SynthRAR and ground truth are summarized in Fig.~\ref{hu_rois}. The zoomed-in profile demonstrates that SynthRAR can correct the errors significantly compared to artifact-free ground truth images.

\begin{table*}[htbp]
\caption{Performance comparisons on MMWHS and RibFrac datasets. The supervised learning methods are trained and tested on the same datasets. The performances are measured by Mean Absolute Error (MAE) in HU, Peak Signal-to-Noise Ratio (PSNR) in dB and Structural Similarity Index Measure (SSIM) by mean (std.).}
\label{supervised_performance}
\centering
\begin{tabular}{clcccccc}
\hline
\multirow{2}{*}{Category}                                                         & \multicolumn{1}{c}{\multirow{2}{*}{Method}} & \multicolumn{3}{c}{MMWHS}                                                     & \multicolumn{3}{c}{RibFrac}                                                   \\ \cline{3-8} 
                                                                                  & \multicolumn{1}{c}{}                        & \multicolumn{1}{c}{MAE} & \multicolumn{1}{c}{SSIM} & \multicolumn{1}{c}{PSNR} & \multicolumn{1}{c}{MAE} & \multicolumn{1}{c}{SSIM} & \multicolumn{1}{c}{PSNR} \\ \hline
\multirow{3}{*}{\begin{tabular}[c]{@{}c@{}}Supervised on \\ real CT\end{tabular}} & AST                                         & 13.1   (2.5)            & 0.937   (0.020)          & 40.66 (1.86)             & 12.6   (2.8)            & 0.936   (0.025)          & 42.89 (2.81)             \\
                                                                                  & DeepRAR                                     & 18.8   (4.0)            & 0.896   (0.029)          & 37.51 (1.95)             & 14.9   (3.0)            & 0.915   (0.030)          & 41.24 (3.39)             \\
                                                                                  & NAFNet                                      & 13.9   (2.6)            & 0.930 (0.026)            & 40.10 (1.79)             & 13.4   (2.9)            & 0.930  (0.026)           & 42.19 (2.59)             \\ \hline
\multirow{4}{*}{Model-based}                                                      & FBP                                         & 233.7 (46.1)            & 0.317   (0.052)          & 12.10 (1.34)             & 171.7 (47.5)            & 0.382   (0.067)          & 14.19 (1.93)             \\
                                                                                  & Norm                                        & 81.9 (9.8)              & 0.566   (0.051)          & 19.20 (0.74)             & 63.8 (10.3)             & 0.611   (0.054)          & 20.35 (1.23)             \\
                                                                                  & WaveFFT                                     & 139.7 (31.4)            & 0.420   (0.074)          & 15.92 (1.48)             & 98.6 (25.2)             & 0.498 (0.084)          & 18.19 (1.68)             \\
                                                                                  & Super                                       & 126.5 (10.3)            & 0.414   (0.033)          & 16.85 (0.64)             & 120.0 (12.0)            & 0.441   (0.036)          & 17.00 (0.80)             \\ \hline
Unsupervised                                                                      & Riner                                       & 42.6   (38.4)           & 0.813   (0.144)          & 28.75 (4.94)             & 31.2   (20.4)           & 0.835   (0.065)          & 28.34 (2.90)             \\ \hline
Ours                                                                              & SynthRAR                                    & 4.3 (1.5)& 0.992   (0.003)          & 47.82 (2.65)             & 4.8 (1.2)& 0.986   (0.005)          & 46.27 (2.23)             \\ \hline
\end{tabular}
\end{table*}

\subsection{Performance comparison to SOTA methods on out-of-distribution dataset}
\subsubsection{RIRE dataset}
The performance of the RAR task on different HU distributions, i.e., brain CT images, are summarized in Table~\ref{generalization_RIRE}. SynthRAR achieves the best performance, as it was trained on synthetic CT-like images without any bias from background information, despite HU distribution shifts. In contrast, the SOTA supervised learning DeepRAR showed significant performance drop due to the large domain gap between train-test pairs (MMWHS–RIRE and RibFrac–RIRE). The state-of-the-art unsupervised Riner with INR performed consistently across different anatomies. However, due to its tendency to overfit individual data points, its expensive time cost limits its clinical applicability. 

\begin{table*}[htbp]
\caption{Performance comparisons on RIRE dataset for HU value out-of-domain validation. The supervised learning methods are trained on the MMWHS and RibFrac datasets separately. The performances are measured by Mean Absolute Error (MAE) in HU, Peak Signal-to-Noise Ratio (PSNR) in dB and Structural Similarity Index Measure (SSIM) by mean (std.).}
\label{generalization_RIRE}
\centering
\begin{tabular}{clcccccc}
\hline
\multirow{2}{*}{Category}                                                         & \multicolumn{1}{c}{\multirow{2}{*}{Method}} & \multicolumn{3}{c}{MMWHS2RIRE}                                                     & \multicolumn{3}{c}{RibFrac2RIRE}                                                            \\ \cline{3-8} 
                                                                                  & \multicolumn{1}{c}{}                        & \multicolumn{1}{c}{MAE} & \multicolumn{1}{c}{SSIM} & \multicolumn{1}{c}{PSNR} & MAE                              & \multicolumn{1}{c}{SSIM} & \multicolumn{1}{c}{PSNR} \\ \hline
\multirow{3}{*}{\begin{tabular}[c]{@{}c@{}}Supervised on \\ real CT\end{tabular}} & AST                                         & 7.3   (0.6)             & 0.985   (0.004)          & 45.78 (1.20)             & \multicolumn{1}{l}{4.7   (0.5)}  & 0.987   (0.004)          & 48.78 (1.57)             \\
                                                                                  & DeepRAR                                     & 31.7   (2.2)            & 0.896   (0.003)          & 35.58 (0.85)             & \multicolumn{1}{l}{24.5   (4.4)} & 0.764   (0.035)          & 36.73 (2.34)             \\
                                                                                  & NAFNet                                      & 5.9   (0.6)             & 0.983   (0.004)          & 45.63 (1.58)             & \multicolumn{1}{l}{4.9   (0.6)}  & 0.986   (0.004)          & 47.90 (1.42)             \\ \hline
\multicolumn{2}{c}{}                                                                                                            & \multicolumn{3}{c}{RIRE}                                                      & \multicolumn{3}{c}{\multirow{7}{*}{-}}                                                 \\ \cline{1-5}
\multirow{4}{*}{Model-based}                                                      & FBP                                         & 106.2 (27.7)            & 0.406   (0.065)          & 16.57 (1.71)             & \multicolumn{3}{c}{}                                                                   \\
                                                                                  & Norm                                        & 40.6 (6.4)              & 0.667   (0.079)          & 23.44 (1.48)             & \multicolumn{3}{c}{}                                                                   \\
                                                                                  & WaveFFT                                     & 61.4 (11.8)             & 0.541  (0.109)          & 20.46 (1.24)             & \multicolumn{3}{c}{}                                                                   \\
                                                                                  & Super                                       & 82.9 (9.0)              & 0.495  (0.071)          & 18.40 (0.70)             & \multicolumn{3}{c}{}                                                                   \\ \cline{1-5}
Unsupervised                                                                      & Riner                                       & 15.1   (16.7)           & 0.906  (0.100)          & 32.06 (4.59)             & \multicolumn{3}{c}{}                                                                   \\ \cline{1-5}
Ours                                                                              & SynthRAR                                    & 2.6   (0.3)             & 0.987   (0.004)          & 50.17 (1.09)             & \multicolumn{3}{c}{}                                                                   \\ \cline{1-5}
\end{tabular}
\end{table*}

\subsubsection{LDCT dataset}
\begin{figure}[htbp]
\centering{\includegraphics[width=8cm]{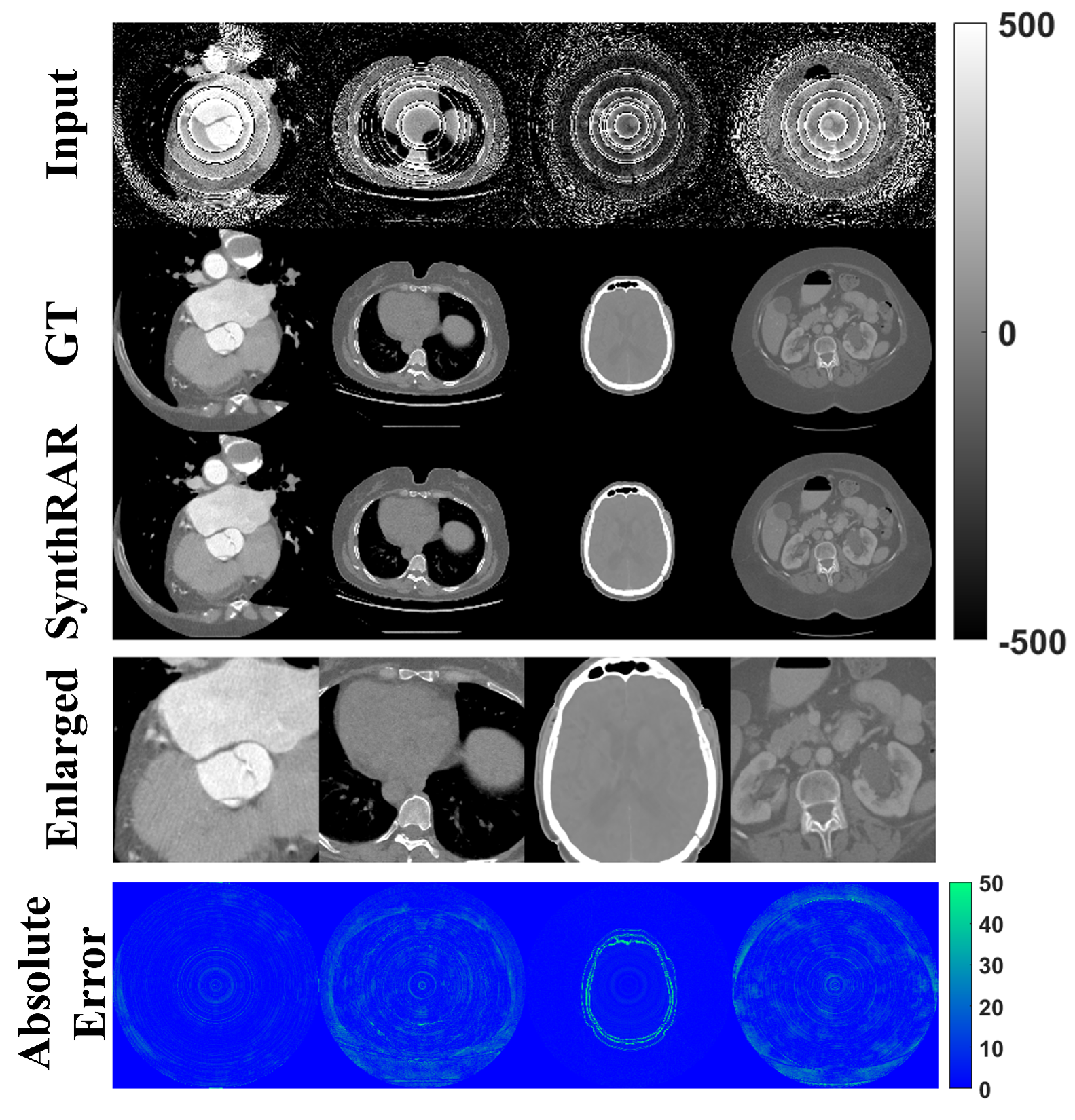}}
\caption{Visual comparisons for 4 different testing datasets. The CT images are having unit of HU values. First column: MMWHS, second column: RibFrac, third column: RIRE, and last colum: LDCT (from left to right).}
\label{fourdata}
\end{figure}

The RAR performances across different scanning geometry, i.e., LDCT images, are summarized in Table~\ref{generalization_LDCT}. The proposed SynthRAR  achieves the best performance, outperforming SOTA supervised learning methods. The supervised DeepRAR trained on real CT datasets, such as MMWHS and RibFrac images, performs worst on LDCT dataset, which contains more background anatomies, such as pelvic regions, making the task more challenging for the model with U-Net architecture. The Riner method also yields poorer results due to HU-value shift, as the prior distribution of true CT values is missing. Qualitative results of the proposed SynthRAR on four different testing datasets are shown in Fig.~\ref{fourdata}. As observed, the out-of-distribution images result in higher absolute errors than the in-distribution ones, i.e., MMWHS and RibFrac images. Overall, the proposed method achieves consistently outperforms SOTA approaches under challenging experimental setups.

\begin{table*}[htbp]
\caption{Performance comparisons on LDCT dataset for geometry out-of-domain validation. The supervised learning methods are trained on the MMWHS and RibFrac datasets separately. The performances are measured by Mean Absolute Error (MAE) in HU, Peak Signal-to-Noise Ratio (PSNR) in dB and Structural Similarity Index Measure (SSIM) by mean (std.).}
\label{generalization_LDCT}
\centering
\begin{tabular}{clcccccc}
\hline
\multirow{2}{*}{Category}                                                         & \multicolumn{1}{c}{\multirow{2}{*}{Method}} & \multicolumn{3}{c}{MMWHS2LDCT}                                                     & \multicolumn{3}{c}{RibFrac2LDCT}                                                                \\ \cline{3-8} 
                                                                                  & \multicolumn{1}{c}{}                        & \multicolumn{1}{c}{MAE} & \multicolumn{1}{c}{SSIM} & \multicolumn{1}{c}{PSNR} & MAE                                  & \multicolumn{1}{c}{SSIM} & \multicolumn{1}{c}{PSNR} \\ \hline
\multirow{3}{*}{\begin{tabular}[c]{@{}c@{}}Supervised on \\ real CT\end{tabular}} & AST                                         & 9.8   (0.8)           & 0.953 (0.006)            & 38.47 (0.90)             & \multicolumn{1}{l}{11.5   (0.8)}   & 0.934 (0.008)            & 37.37 (0.70)             \\
                                                                                  & DeepRAR                                     & 217.3   (21.9)        & 0.350 (0.079)            & 14.41 (0.55)             & \multicolumn{1}{l}{145.9   (14.2)} & 0.630 (0.040)            & 17.94 (0.57)             \\
                                                                                  & NAFNet                                      & 9.8   (0.7)           & 0.948 (0.006)            & 38.55 (0.84)             & \multicolumn{1}{l}{10.5   (0.8)}   & 0.953 (0.006)            & 38.35 (0.67)             \\ \hline
\multicolumn{2}{c}{}                                                                                                            & \multicolumn{3}{c}{LDCT}                                                      & \multicolumn{3}{c}{\multirow{7}{*}{-}}                                                     \\ \cline{1-5}
\multirow{4}{*}{Model-based}                                                      & FBP                                         & 189.4   (20.2)        & 0.316 (0.030)            & 13.62 (0.67)             & \multicolumn{3}{c}{}                                                                       \\
                                                                                  & Norm                                        & 51.2   (8.4)          & 0.655 (0.051)            & 22.19 (1.73)             & \multicolumn{3}{c}{}                                                                       \\
                                                                                  & WaveFFT                                     & 64.3   (4.9)          & 0.582 (0.037)            & 21.42 (0.73)             & \multicolumn{3}{c}{}                                                                       \\
                                                                                  & Super                                       & 83.4   (13.9)         & 0.511 (0.062)            & 19.78 (1.60)             & \multicolumn{3}{c}{}                                                                       \\ \cline{1-5}
Unsupervised                                                                      & Riner                                       & 62.3 (4.8)            & 0.912 (0.010)            & 25.19 (0.40)             & \multicolumn{3}{c}{}                                                                       \\ \cline{1-5}
Ours                                                                              & SynthRAR                                    & 4.7 (0.6)           & 0.995 (0.001)            & 45.93 (1.32)             & \multicolumn{3}{c}{}                                                                       \\ \cline{1-5}
\end{tabular}
\end{table*}

\subsection{Ablation study on estimation networks}
We introduced networks to estimate the IR and IM data in the sinogram, which are used to correct the non-ideal detector response in the forward projection operation $\tilde{A}x = (-\ln{{H}_{IR}} + Ax)\times{H}_{IM}$. The effectiveness of these two networks is summarized in Table~\ref{ablation_IR_IM}. Four different cases are considered in the ablation studies.

\begin{figure}[htbp]
\centering{\includegraphics[width=6cm]{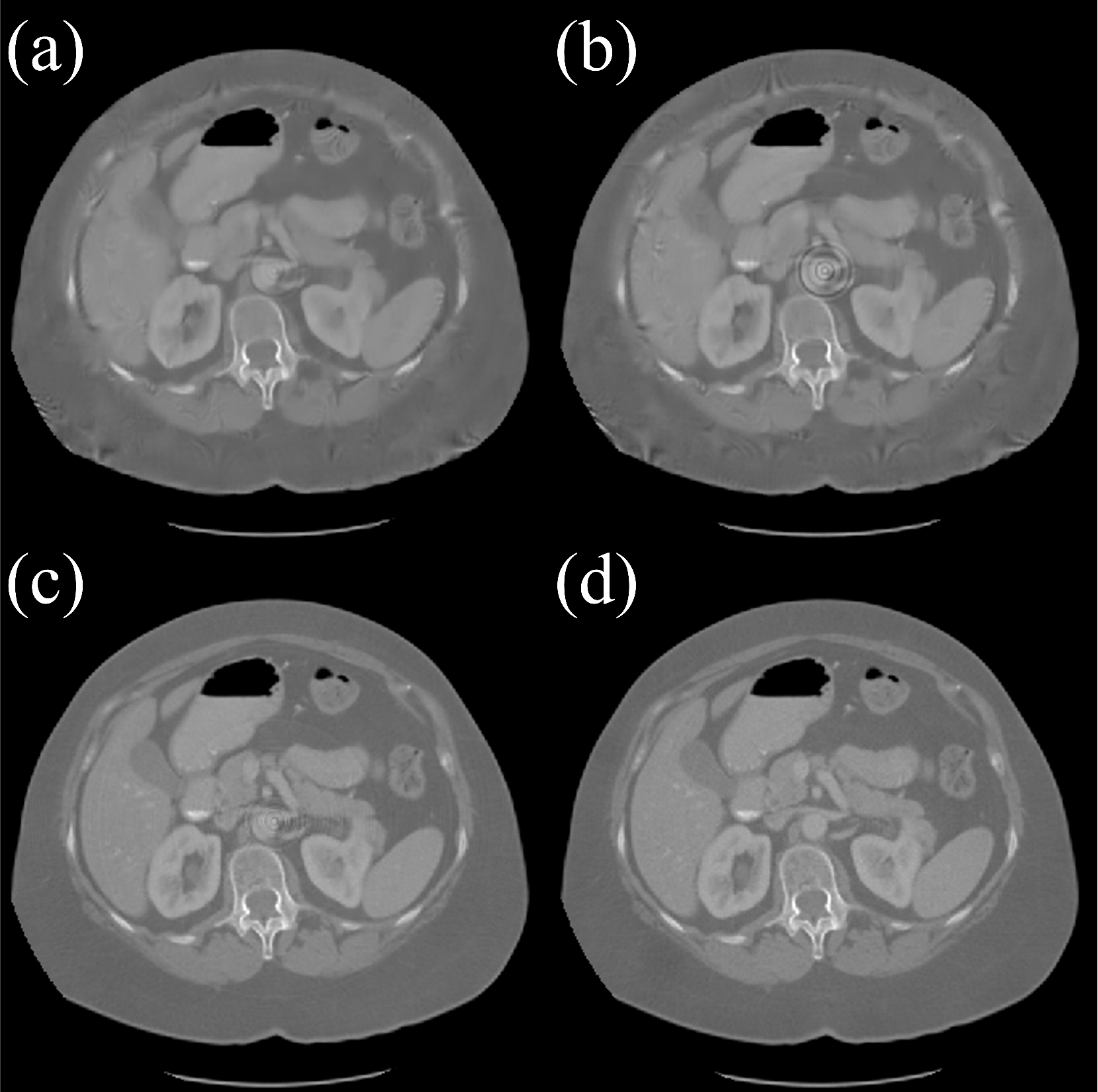}}
\caption{Qualitative visualization of the impact for different modules, example from LDCT dataset. (a) Backbone w/o IR and IM, (b) Backbone w/o IM, (c) Backbone w/o IR, and (d) Full model.}
\label{ablation}
\end{figure}

\begin{figure}[htbp]
\centering{\includegraphics[width=6cm]{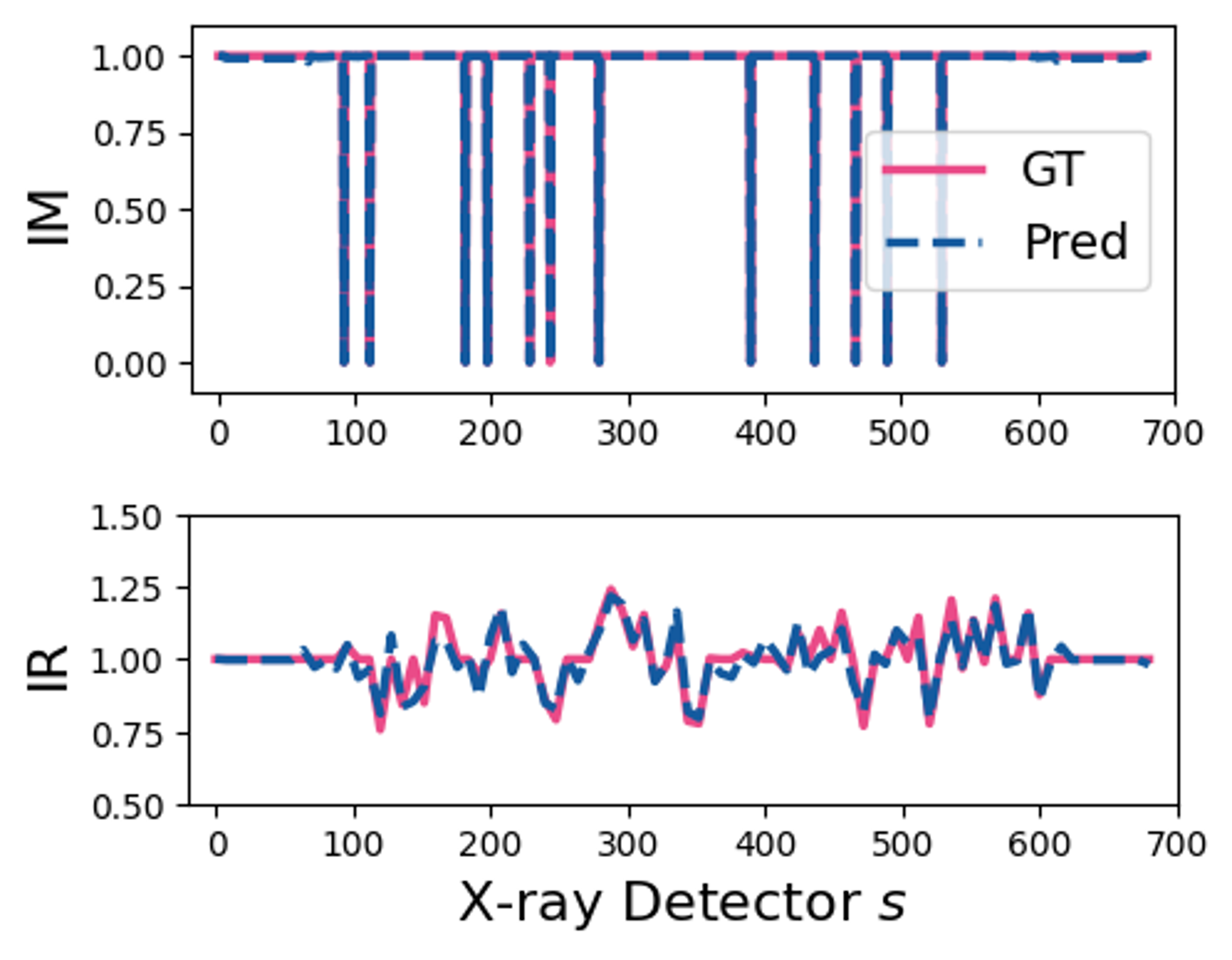}}
\caption{Parameter estimations of SynthRAR modules for estimating IR and IM detectors.}
\label{ablation_curve}
\end{figure}

\subsubsection{Backbone ISTA-Net}
The backbone ISTA-Net is applied to the considered task. There are no modules to correct IR and IM points in the sinogram, simple CNN filters are applied instead. The forward projection is simplified as $\tilde{A}x = Ax$.
\subsubsection{SynthRAR w/o IM}
The proposed model without IM estimation net, which therefore no estimation of the dead detectors but just bias from inconsistent responses. The forward projection is formulated as $\tilde{A}x = -\ln{{H}_{IR}} + Ax$.
\subsubsection{SynthRAR w/o IR}
The projected model without IR estimation net, which is able to estimate the dead detectors, but cannot correct the energy bias due to IR points. The forward projection is formulated as $\tilde{A}x = {H}_{IM}\times{Ax}$.
\subsubsection{SynthRAR}
The full model with all modules to estimate IR and IM with forward projection $\tilde{A}x = (-\ln{{H}_{IR}} + Ax)\times{H}_{IM}$.

\begin{table}[htbp]
\centering
\caption{Ablation of the impacts of different estimation nets. The performances are measured by Mean Absolute Error (MAE) in HU, Peak Signal-to-Noise Ratio (PSNR) in dB and Structural Similarity Index Measure (SSIM) by mean (std.).}
\label{ablation_IR_IM}
\begin{tabular}{ccccc}
\hline
\multicolumn{2}{c}{Estimation Net}                    & \multicolumn{3}{c}{MMWHS}                                                     \\ \hline
IR                        & IM                        & \multicolumn{1}{c}{MAE} & \multicolumn{1}{c}{SSIM} & \multicolumn{1}{c}{PSNR} \\ \hline
$\times$                  & $\times$                  & 20.2   (3.4)            & 0.876 (0.033)            & 33.69 (1.34)             \\
\checkmark & $\times$                  & 20.3   (3.5)            & 0.875  (0.032)
            & 33.60 (1.34)             \\
$\times$                  & \checkmark & 5.8   (1.3)             & 0.985  (0.005)
            & 44.87 (1.67)             \\
\checkmark & \checkmark & 4.3   (1.5)             & 0.992  (0.003)
            & 47.82 (2.65)             \\ \hline
\multicolumn{2}{c}{Estimation Net}                    & \multicolumn{3}{c}{RibFrac}                                                   \\ \hline
IR                        & IM                        & \multicolumn{1}{c}{MAE} & \multicolumn{1}{c}{SSIM} & \multicolumn{1}{c}{PSNR} \\ \hline
$\times$                  & $\times$                  & 18.7   (2.5)            & 0.875  (0.029)            & 33.69 (1.12)             \\
\checkmark & $\times$                  & 18.3   (2.8)            & 0.872  (0.031)
            & 33.85 (1.23)             \\
$\times$                  & \checkmark & 6.9   (1.2)             & 0.974  (0.008)
            & 43.15 (1.55)             \\
\checkmark & \checkmark & 4.8   (1.2)             &  0.986  (0.005)
            & 46.27 (2.23)             \\ \hline
\multicolumn{2}{c}{Estimation Net}                    & \multicolumn{3}{c}{RIRE}                                                      \\ \hline
IR                        & IM                        & \multicolumn{1}{c}{MAE} & \multicolumn{1}{c}{SSIM} & \multicolumn{1}{c}{PSNR} \\ \hline
$\times$                  & $\times$                  & 6.2   (1.0)             & 0.911 (0.039)   & 40.19 (1.18)             \\
\checkmark & $\times$                  & 6.3   (0.8)             & 0.914  (0.033)  & 40.03 (1.10)             \\
$\times$                  & \checkmark & 3.2   (0.3)             & 0.981  (0.006)
            & 47.90 (1.48)             \\
\checkmark & \checkmark & 2.6   (0.3)             & 0.987  (0.004)       & 50.17 (1.09)             \\ \hline
\multicolumn{2}{c}{Estimation Net}                    & \multicolumn{3}{c}{LDCT}                                                      \\ \hline
IR                        & IM                        & \multicolumn{1}{c}{MAE} & \multicolumn{1}{c}{SSIM} & \multicolumn{1}{c}{PSNR} \\ \hline
$\times$                  & $\times$                  &       12.7 (0.8)      &  0.946 (0.007)  &       36.32 (0.93)      \\
\checkmark & $\times$                  &       12.1 (0.9)       & 0.942 (0.008) &     36.42 (0.98)        \\
$\times$                  & \checkmark &       6.2 (0.6)      &  0.988 (0.002)  &       43.51 (1.05)      \\
\checkmark & \checkmark &      4.7 (0.6)        &    0.995 (0.001)   &     45.93 (1.32)        \\ \hline
\end{tabular}
\end{table}

From the results, the naive ISTA-Net cannot correct any ring artifacts due to lack of LPP pattern estimation, but instead introducing blurring. By gradually introducing the estimation networks, the model gains ability to compensate the LPP information, and is able to correct the ring artifacts in different definitions. More specifically, invalid measurement leads to higher image quality degradation due to its strong impact of the ring. Proper modulation of IM can drastically improve the final image quality. In contrast, the IR points have lower impacts on the output image quality, this also indicates the IR effects in CT imaging are less severe for image-based diagnose. By jointly considering the two modules with ISTA-Net, the trained model can consistently perform the best in on four datasets. The qualitative comparisons are shown in Fig.~\ref{ablation}, and one example of parameter estimations for IR and IM detectors are shown in Fig.~\ref{ablation_curve}. 
\subsection{Ablation study of training on real CT}
Instead of training SynthRAR on synthetic CT-like images, we conducted an ablation study to evaluate the impact of synthetic images on performance. Specifically, two different models were trained on MMWHS and RibFrac datasets, denoted MMWHS-RAR and RibFrac-RAR. Each model trained on CT images was evaluated on its own data domain and further tested on RIRE and LDCT datasets to assess generalization. The performance results are summarized in Table~\ref{on_real}. 

Comparing to SynthRAR model, real-CT image trained models performs better on their own data domain, i.e., MMWHS-RAR got better performance than SynthRAR on MMWHS images, while RibFrac-RAR got better performance than SynthRAR on RibFrac images. However, when evaluating their generalizations, SynthRAR achieves better results on RIRE dataset, while other two models trained on specific anatomies are degraded. On both MMWHS and RibFrac dataset, SynthRAR performs similar to the domain-specific models, which indicates our synthetic data strategy can provide a generalized solution without requirement of the real-world images for training. Therefore, data collection efforts for model training are significantly reduced.

\begin{table}[htbp]
\centering
\caption{Comparison of the proposed model trained on real CT dataset. The performances are measured by Mean Absolute Error (MAE) in HU, Peak Signal-to-Noise Ratio (PSNR) in dB and Structural Similarity Index Measure (SSIM) by mean (std.).}
\label{on_real}
\begin{tabular}{cccc}
\hline
MMWHS-RAR   & MAE         & SSIM           & PSNR         \\ \hline
MMWHS       & 3.9 (1.2) & 0.995 (0.002)& 48.74 (2.32) \\
RibFrac     & 8.8 (3.8) & 0.941 (0.047)& 41.80 (3.80) \\
RIRE        & 5.5 (0.9) & 0.979 (0.003)& 47.24 (1.95) \\
 LDCT& 15.5 (0.9)& 0.920 (0.009)&35.66 (0.65)
\\ \hline
RibFrac-RAR & MAE       & SSIM           & PSNR         \\ \hline
MMWHS       & 9.3 (2.2) & 0.985 (0.009)& 43.10 (3.21) \\
RibFrac     & 4.4 (1.1) & 0.989 (0.005)& 46.96 (2.03) \\
RIRE        & 4.5 (0.6) & 0.977 (0.048)& 46.74 (1.18) \\
LDCT& 14.1 (0.5)& 0.892 (0.012)&36.31 (0.80)\\ \hline
SynthRAR    & MAE       & SSIM           & PSNR         \\ \hline
MMWHS       & 4.3 (1.5) & 0.992 (0.003)& 47.82 (2.65) \\
RibFrac     & 4.8 (1.2) &  0.986 (0.005)& 46.27 (2.23) \\
RIRE        & 2.6 (0.3) & 0.987 (0.004)& 50.17 (1.09) \\
LDCT& 4.7 (0.6)& 0.995 (0.001)&45.93 (1.32)
\\ \hline
\end{tabular}
\end{table}

\subsection{Computational cost}
The computational costs of deep learning based methods are evaluated in Tabel.~\ref{time_cost}. As for supervised methods, DeepRAR is most cheap one as the simplicity of UNet design, while AST and NAFNet are complex in time due to design of complexity modules in the network, However NAFNet is designed to have lower GPU cost. The proposed SynthRAR is the most expensive one with time cost of 110 ms with higher GPU consumption, which is due to iterative stages of ISTA-Net and forward project in CT operations in whole image level. However, compared to online optimized Riner with the lowest memory cost, the Riner not feasible to be deployed to on-device situation. Although the proposed SynthRAR is expensive than other supervised methods, it is still suitable for edge deployment with acceptable cost and image quality rewards.

\begin{table}[htbp]
\centering
\caption{Computational costs for different deep learning methods. Time is measured in millisecond (ms) or minute (min.), while the peak GPU memory allocation is measured in Gigabytes (GB) on A100.}
\label{time_cost}
\begin{tabular}{ccc}
\hline
Methods           & Time & RAM (GB)   \\ \hline
AST           &  66 ms & 2.07  \\ 
DeepRAR       & 2 ms & 0.24\\
NAFNet       & 23 ms & 0.34\\
Riner        & $>$ 10 min.& 0.11\\
SynthRAR    & 110 ms& 2.45
\\ \hline
\end{tabular}
\end{table}

\section{Conclusions}\label{conclusions}
In this paper, a synthetic-data-driven dual-domain LPP correction method was proposed, combining the estimation of inconsistent response and invalid measurement detectors via CNNs. Unlike existing supervised RAR methods, a novel synthetic data generation strategy was introduced to mitigate data collection costs for model training, while integrating physical background knowledge from LPPs. Moreover, based on the definition of LPP, failure detectors were estimated from observed sinogram images using designed estimation networks, which can guide the ISTA-Net to correct the data under the modified forward projection. The jointly trained model on synthetic CT-like images shows promising performance on challenging testing cases on four different public datasets, and outperforms the-state-of-the-art methods.

Although the proposed method provides promising correction for CT with ring artifacts, its feasibility on multi-row fan-beam CT or cone-beam CT requires further study and validation. In addition, in state-of-the-art CT scanners designed for low-dose and fast imaging, short scanning is enabled with rotation angles less than $360^\circ$, which is commonly reconstructed with advanced algorithms rather than standard filtered back projection. Advanced implementation of this type of method in deep learning-enabled solution is required for next-generation CT scanners. More importantly, in common industrial practice, vendors are very strict to the number of IR and IM points in the detector, as high-fidelity image is critical for clinical applications. Therefore, the experimental assumptions of SynthRAR is too aggressive, and vendors would like to discard the detector directly in such situation. Future study should focus on the on-device evaluation using data from scanners with low performing pixels, which will require great efforts in collaboration between researchers and engineers. Therefore, further adjustment, or even testing-time adaption methods should be considered for each different scanners.
\bibliographystyle{IEEEtran}
\bibliography{draft}

\end{document}